# Generative AI for Named Entity Recognition in Low-Resource Language Nepali

**Sameer Neupane, Jeevan Chapagain**
University of Memphis
Memphis, Tennessee, USA
sameerthereds@gmail.com, jchpgain@memphis.edu

**Nobal B. Niraula, Diwa Koirala**
Nowa Lab
Madison, Alabama, USA
nobal@nowalab.com, diwa@nowalab.com

## Abstract

Generative Artificial Intelligence (GenAI), particularly Large Language Models (LLMs), has significantly advanced Natural Language Processing (NLP) tasks, such as Named Entity Recognition (NER), which involves identifying entities like person, location, and organization names in text. LLMs are especially promising for low-resource languages due to their ability to learn from limited data. However, the performance of GenAI models for Nepali, a low-resource language, has not been thoroughly evaluated. This paper investigates the application of state-of-the-art LLMs for Nepali NER, conducting experiments with various prompting techniques to assess their effectiveness. Our results provide insights into the challenges and opportunities of using LLMs for NER in low-resource settings and offer valuable contributions to the advancement of NLP research in languages like Nepali.

## Introduction

Generative Large Language Models (LLMs) like GPTs (Generative Pretrained Transformers) are prominent examples of Generative AI (GenAI). These LLMs play a significant role in advancing Natural Language Processing (NLP) by enabling machines to understand, generate, and manipulate human language in more sophisticated and flexible ways. They have enabled several NLP applications such as dialog systems (Roller et al. 2021), text generation (Brown et al. 2020), machine translation (Wu et al. 2016), text summarization (Zhang et al. 2020), question answering (Raffel et al. 2020), text-based reasoning (Wei et al. 2022), and information extraction (Wang et al. 2023; Petroni et al. 2019).

LLMs like GPT4 (Achiam et al. 2023) are highly effective few-shot learners, meaning they can perform tasks with minimal task-specific examples. Unlike traditional machine learning models that require large amounts of labeled training data, LLMs leverage their vast pre-training on diverse multilingual corpora to understand patterns, syntax, and semantics in language. When given a few examples of a task in a prompt (known as few-shot learning), the model can generalize from those examples and apply the learned patterns to new, unseen inputs. These abilities are particularly interesting for low-resource languages like Nepali where manually compiling extensive datasets can be expensive and time-consuming. This ability opens up new possibilities for expanding NLP capabilities in Nepali, enhancing accessibility, and improving the effectiveness of AI-driven language tools without the need for large-scale, language-specific data.

The effectiveness of LLMs typically depends on the availability of vast amounts of training data during pre-training. However, for low-resource languages like Nepali, the volume of such data is considerably smaller compared to resource-rich languages like English, raising questions about their performances. This paper investigates the performance of state-of-the-art LLMs in processing Nepali language, specifically focusing on their Named Entity Recognition (NER) capabilities.

NER is a key task in NLP that involves identifying and classifying named entities in text into predefined categories. The named entities could be Person, Location, Organization, and so on. By automatically detecting and labeling these named entities, NER systems help extract structured information from free-form text, enabling more effective analysis and understanding of the content. NER serves as a crucial component in various downstream applications like question answering (Mollá, Van Zaanen, and Smith 2006), machine translation (Babych and Hartley 2003), and text summarization (Li et al. 2020).

Our specific contributions in this paper include:

- Benchmarking the state-of-the-art Generative LLM for NER in low-resource language Nepali
- Exploring effective prompting techniques for NER
- Investigating the impact of prompt translation
- Discussing practical considerations for leveraging LLMs in low-resource language settings

## Related Works

NER has been extensively studied in NLP, with English and German languages receiving the most attention. However, research on NER for the Nepali language remains limited. This section presents an overview of the existing work in Nepali NER and overall in NER using LLMs.

## NER in Nepali Language

One of the earliest contributions to Nepali NER was made by Bam et al. (Bam and Shahi 2014) who proposed a Support Vector Machine (SVM) architecture for classifying named entities into four categories: Person, Location, Organization, and Miscellaneous. Their approach used a one-versus-rest classification setting. Dey and Prukayastha (Dey, Paul, and Purkayastha 2014) employed a Hidden Markov Model with n-gram techniques to extract Part-of-Speech (POS) tags. Their method combined POS tags containing common nouns, proper nouns, or a combination of both, and then used a gazetteer list as a look-up table to identify named entities. Singh et al. (Singh, Padia, and Joshi 2019) focused on Person, Location, and Organization entities. The authors trained various neural models with different word embeddings. Their findings indicated that the BiLSTM-CNN model performed the best among the tested architectures.

Recently, Niraula and Chapagain conducted a detailed study of NER in Nepali news (Niraula and Chapagain 2022) and Nepali Tweets (Niraula and Chapagain 2023). They expanded the entity types to five categories: Person, Location, Organization, Event, and Date. The authors provided comprehensive annotation guidelines for labeling each entity type as well as publishing standard datasets for training and testing. Notably, their research demonstrated that transformer models can achieve state-of-the-art performance for resource-poor languages like Nepali. Building on this work, Subedi et al.(2024) evaluated BERT variants for Nepali NER, showing NepBERTa outperformed multilingual alternatives while traditional approaches like BiLSTM+CNN and SVM remained competitive, demonstrating the value of both neural and classical methods.

## NER using LLMs

The emergence of LLMs has introduced new possibilities for NER tasks across various languages. He et al. (He et al. 2023) introduced a novel prompt-based contrastive learning method that addresses the limitations of traditional N-gram traversal approaches in few-shot NER. Their method eliminates the need for template construction and label word mappings by leveraging external knowledge to initialize semantic anchors for each entity type. These anchors, when appended with input sentence embeddings as template-free prompts (TFPs), enable efficient entity recognition without extensive template engineering.

A significant advancement in bridging the gap between LLMs and NER was introduced by Wang et al.(2023) addressing the fundamental mismatch between NER's sequence labeling nature and LLMs' text generation capabilities. Their GPT-NER framework innovatively transforms the sequence labeling task into a generation task by introducing special token markers (e.g., "@@Columbus##" for location entities). To combat LLMs' tendency to hallucinate entities when processing null inputs, they developed a self-verification strategy where the model validates its entity classifications. This approach significantly narrowed the performance gap between LLM-based and supervised NER systems, particularly benefiting low-resource scenarios where traditional supervised baselines often struggle.

To improve few-shot NER, Ye et al.(2024) introduced LLM-DA, a data augmentation technique that leverages LLMs' rewriting capabilities and world knowledge. Their method combines contextual rewriting strategies with entity-level augmentation, addressing the semantic integrity issues common in traditional data augmentation approaches. With different contextual rewriting strategies and strategic noise injection, LLM-DA demonstrated superior data quality and enhanced model performance.

While these advances in LLM-based NER have shown promising results, their application to the Nepali language remains unexplored. The success of LLMs in addressing NER tasks for low-resource languages is yet to be studied in Nepali NER, particularly through techniques such as few-shot learning and prompt engineering. This research gap presents an opportunity for us to investigate how LLMs can improve the performance of Nepali NER.

## Methodology

We adopted the methodology from Wang et al. (2023) for our NER tasks. Our overall architecture is summarized in Figure 1. The process starts with selecting the base prompt. We construct the final prompt by extending the base prompt with the examples obtained via various selection strategies and use that prompt to predict the output using a LLM. We also have the self-verification step to further improve the quality of the predictions. We describe these components below.

### Prompts for NER

We developed base prompts in English and Nepali languages based on the foundational approach outlined in (Wang et al. 2023). This process involved constructing prompts that instructed the LLM to detect and label specific entity types using predefined delimiters (@@ and ##). In few-shot settings, we formatted examples using these delimiters to enclose target entities, maintaining consistency with the methodology described in the original paper. However, the exact prompts provided in the paper did not yield consistent results for Nepali sentences. To address this, we engaged in prompt engineering, refining, and iterative modification testing to adapt the prompts to Nepali language contexts. The finalized prompts both in English and Nepali are shown in Table 1 and Table2. In the examples, prompts are used to identify ORGANIZATION entities (युएइ (UAE) and नेपाल (Nepal)) in the given test sentence.

We parse the LLM output and convert it to the standard BIO format (Beginning-Inside-Outside) for NER using the entity delimiters.

### Sampling Examples for Zero and Few-shot Settings

For zero-shot prompting, no examples are included in the base prompt. This approach relied solely on the model's inherent ability to understand the task based on the instructions provided. To ensure the output adhered to the desired and consistent format, we appended the prompt with: *Output the whole sentence and enclose the entity within @@ and ##.* This explicit directive helped guide the model to generate

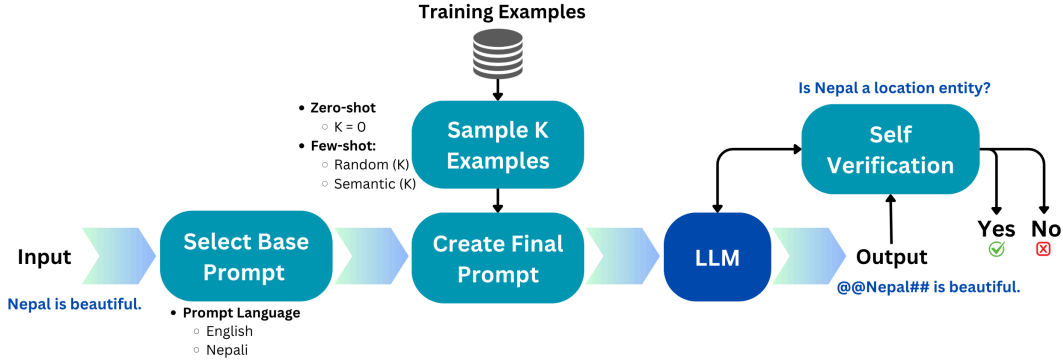

Figure 1: Overall Architecture

| Base Prompt | The task is to label Organization entities in the given Nepali sentence. Below are some examples with Input and Output pairs. For the prediction, you should generate the output in the same format as in the examples. Do not give any explanations. |
|---|---|
| Examples | **Input:** तर, भारत को राष्ट्रिय टोली बाट खेली सकेका खेलाडी लाई बिसिसिआई ले अन्य देश को लागि मा खेल्ने अनुमति दिँदैन। (*However, BCCI does not allow players who have already played for the Indian national team to play for other countries.*) <br> **Output:** तर, @@भारत## को राष्ट्रिय टोली बाट खेली सकेका खेलाडी लाई @@बिसिसिआई## ले अन्य देश को लागि मा खेल्ने अनुमति दिँदैन। |
| Test Instance | Now predict the output for the following sentence. <br> **Input:** अंक तालिका मा युएइ शीर्षस्थान मा छ भने नेपाल दोस्रो स्थान मा छ। (*In the points table, UAE is at the top position while Nepal is in second position.*) |
| Output | अंक तालिका मा @@युएइ## शीर्षस्थान मा छ भने @@नेपाल## दोस्रो स्थान मा छ। |

Table 1: English prompt for ORGANIZATION with one example and its output for a test sentence. Content inside () is the English translation and is not the part of the final prompt.

| Base Prompt | गरिनुपर्ने काम भनेको दिइएको नेपाली वाक्यमा सङ्घ संस्थाको नामलाई @@ ## भित्र लेबल गर्नु हो। तल वाक्यलाई लेबल गर्दा उदाहरणको जस्तै ढाँचामा मात्र गर्नुहोस्। कुनै थप व्याख्या नगर्नुहोस्। (*The task is to label Organization entities in the given Nepali sentence within @@ ##. Below are some examples with Input and Output pairs. For the prediction, you should generate the output in the same format as in the examples. Do not give any explanations.*) |
|---|---|
| Examples | **उदाहरणहरू:** (*Examples:*) <br> **वाक्य:** (*Input:*) तर, भारत को राष्ट्रिय टोली बाट खेली सकेका खेलाडी लाई बिसिसिआई ले अन्य देश को लागि मा खेल्ने अनुमति दिँदैन। <br> **नतिजा:** (*Output:*) तर, @@भारत## को राष्ट्रिय टोली बाट खेली सकेका खेलाडी लाई @@बिसिसिआई## ले अन्य देश को लागि मा खेल्ने अनुमति दिँदैन। |
| Test Instance | अब तल दिइएको वाक्यलाई लेबल गर्नुहोस्। (*Now predict the output for the following sentence.*) <br> **वाक्य:** (*Input:*) अंक तालिका मा युएइ शीर्षस्थान मा छ भने नेपाल दोस्रो स्थान मा छ। |
| Output | अंक तालिका मा @@युएइ## शीर्षस्थान मा छ भने @@नेपाल## दोस्रो स्थान मा छ। |

Table 2: Nepali prompt for ORGANIZATION with one example and its output for a test sentence. Content inside () is the English translation and is not the part of the final prompt.

outputs in the specified format, enhancing clarity and consistency in the results.

For few-shot prompting, we followed the same methodology described in (Wang et al. 2023), which provides a structured framework for incorporating examples into prompts to guide the language model's predictions. Specifically, we used two strategies for selecting $k$ examples: random selection and semantic selection.

**Random Selection** For random selection, we randomly sampled $k$ examples from the training set corresponding to the desired entity type. The prompt then included these examples to demonstrate the expected format and guide the language model's predictions. This method provides a straightforward way to include diverse examples but does not ensure semantic similarity to the test instance.

**Semantic Selection** For the semantic selection, we adopted the embedding strategy using (k-nearest neighbors (k-NN) proposed in (Wang et al. 2023). To implement this, we utilized NPVec1 BERT(Koirala and Niraula 2021) to generate the embeddings of the examples in the training set. NPVec1 BERT, a model fine-tuned on Nepali text, was chosen to ensure the embeddings effectively captured Nepali's linguistic nuances and semantic relationships. We computed the cosine similarity between the embedding of the test in-stance and the embeddings in the training set. The top *k* nearest neighbors were selected based on the same entity type as that of test instance and similarity, ensuring contextually relevant examples to enhance the model's generalization.

## The LLM Model

We considered several open-source and proprietary LLMs including LLAMA (*Llama-3-8B*)[1], Mistral (*Mistral Large*)[2] and GPT4o[3] for our experiments. These models' advanced multilingual capabilities make them particularly suitable for low-resource languages, as they can be adapted to specific tasks through prompts rather than requiring extensive training data and fine-tuning like traditional NER models.

**Output Inconsistencies**   With the same prompt, *Llama-3-8B* and *Mistral Large* models generated very inconsistent output compared to GPT4o, as demonstrated in Table 3. Consistent output is critical for parsing and consuming the output. This inconsistency highlights the need for further prompt engineering or potentially fine-tuning these smaller LLMs to produce reliable outputs. Although some outputs from the GPT4o model required minimal post-processing, they were consistent enough to conduct evaluations effectively. For that reason, we chose GPT4o as our LLM for the experiments. GPT4o is also the state-of-the-art multilingual generative LLM at present.

| |
|---|
| **Input:** नेपाल स्टक एक्सचेन्ज को कारोबार परिपाटी अनुसार यो कारोबार राफसाफ हुन ३ दिन लाग्छ र बिहीबार मात्र यस को भुक्तानी प्राप्त हुन्छ । (*According to the Nepal Stock Exchange's trading protocol, it takes 3 days for this transaction to clear, and the payment will only be received on Thursday*) |
| **LLAMA output:** |
| @@नेपाल स्टक एक्सचेन्ज## कारोबार परिपाटी## |
| **Mistral output:** |
| In this sentence, "स्टक एक्सचेन्ज" (*Stock Exchange*) is identified as an Organization entity and is labeled accordingly. |
| **GPT4o output:** |
| नेपाल @@स्टक एक्सचेन्ज## को कारोबार परिपाटी अनुसार यो कारोबार राफसाफ हुन ३ दिन लाग्छ र बिहीबार मात्र यस को भुक्तानी प्राप्त हुन्छ । |

Table 3: Outputs from different LLMs for the same prompt and input sentence.

## Self-verification

We adopted a self-verification mechanism inspired by (Wang et al. 2023). In this approach, each predicted entity undergoes a secondary validation step using the LLM itself. The model is prompted to confirm whether the identified entity is valid. Predictions that are validated by the model are retained, while those deemed invalid are discarded. This method aims to improve precision by filtering out false positives while preserving valid predictions.

---

[1] https://github.com/meta-llama/llama3
[2] https://mistral.ai/news/mistral-large-2407/
[3] https://openai.com/index/gpt-4/

## Benchmark Data Sets

This study uses EverestNER (Niraula and Chapagain 2022), the largest Nepali NER benchmark datasets for news articles and is annotated with five entity types: LOCATION, ORGANIZATION, PERSON, DATE, and EVENT. The training set includes 847 articles and the test set comprises 149 articles, with Person names being the most frequent (8,822) and Events the least frequent (371) among the 24,587 annotated entities. Table 4 presents the complete dataset statistics.

## Evaluation Metrics

We evaluated the model's performance on EverestNER test set (Niraula and Chapagain 2022) using CoNLL evaluation metrics (Sang and De Meulder 2003), reporting precision, recall, and F1-scores (both micro and macro averages).

# Experiments and Results

To evaluate the results for zero and few shots along with random and semantic selections, we implemented two strategies: a) merging the entities (Merged) and b) without merging (Entity-wise).

**Merged Approach**   In the merging approach, we process the predictions for each of the five entity types and then combine them token-by-token into BIO format. Each entity type is assigned its respective labels (B-Entity, I-Entity, or O) based on the model's prediction for each token in the sentence. During merging, when a token was assigned to multiple entity types, we used a predefined preference order to resolve the conflicts. The priority order is as follows: LOCATION, ORGANIZATION, PERSON, DATE, and EVENT. This means that if a token is predicted as LOCATION and ORGANIZATION, it will be assigned the LOCATION label. This merged prediction is then compared against a ground truth sequence, which includes all entity types, for evaluation purposes.

**Entity-wise (Non-Merged) Approach**   In the entity-wise approach, each entity type is evaluated independently against its corresponding ground truth. Instead of combining all entity types into a single sequence, this approach treats each entity type—LOCATION, ORGANIZATION, PERSON, DATE, and EVENT—separately. The predictions for each entity type are compared with the ground truth sequences for that specific entity type. This allows for a focused evaluation of the model's performance on each entity category without combining them, ensuring that each entity type is assessed in isolation.

## Impact of Number of Shots (Examples)

We investigated the impact of different shot selection strategies (k = 0, 5, and 10) using two selection methods: Random Selection and Semantic Selection as described in the methodology section. Table 5 (without merging) shows the impact of shot selection strategies on NER performance. The baseline model (k=0) achieved an F1-micro score of 0.31, with low precision (0.21) but moderate recall (0.62). Few-shot examples, particularly with random selection, improved performance, with k=5 and k=10 achieving F1-micro scores of 0.52 and 0.55, respectively, and increasing precision and recall. However, semantic selection outperformed both, reaching the highest F1-micro score of 0.64 with k=10, showing improvements in precision (0.51) and recall (0.86).

| Data | Articles | Sentences | Tokens | Avg. Sent. Len | LOC | ORG | PER | EVT | DAT |
|---|---|---|---|---|---|---|---|---|---|
| Train | 847 | 13,848 | 268,741 | 19.40 | 5,148 | 4,756 | 7,707 | 312 | 3,394 |
| Test | 149 | 1,950 | 39,612 | 20.31 | 809 | 715 | 1,115 | 59 | 572 |
| Total | 996 | 15,798 | 308,353 | 19.51 | 5,957 | 5,471 | 8,822 | 371 | 3,966 |

Table 4: EverestNER Data Sets for Nepali News

| Sel. | K | Without Merging | | | | Merging | | | |
|---|---|---|---|---|---|---|---|---|---|
| | | Pr. | Re. | $F_m$ | $F_\mu$ | Pr. | Re. | $F_m$ | $F_\mu$ |
| Base | 0 | 0.21 | 0.62 | 0.31 | 0.31 | 0.17 | 0.43 | 0.23 | 0.24 |
| Rand | 1 | 0.36 | 0.77 | 0.48 | 0.49 | 0.39 | 0.66 | 0.45 | 0.49 |
| | 5 | 0.38 | 0.84 | 0.48 | 0.52 | 0.34 | 0.55 | 0.37 | 0.42 |
| | 10 | 0.40 | 0.87 | 0.50 | 0.55 | 0.38 | 0.62 | 0.41 | 0.47 |
| Sem | 1 | 0.40 | 0.77 | 0.51 | 0.53 | 0.39 | 0.65 | 0.46 | 0.49 |
| | 5 | 0.49 | 0.85 | 0.57 | 0.62 | 0.50 | 0.73 | 0.52 | 0.60 |
| | 10 | 0.51 | **0.86** | 0.57 | **0.64** | 0.53 | **0.75** | 0.54 | **0.63** |

Table 5: NER Results: Merged vs Non-merged Settings. *Note*: Ran, Sem, Pr., Re., $F_m$ and $F_\mu$ denote Random, Semantic Selection, Precision, Recall, F1-macro, and F1-micro, respectively.

Table 5 shows that entity merging reduced model performance, with the baseline F1-micro score dropping to 0.24 and recall declining from 0.62 to 0.43, indicating a significant loss of valid predictions. Although semantic selection with k=10 produced the best merged results (F1-micro: 0.63, F1-macro: 0.54), they were still lower than the non-merged approach, supporting the hypothesis that merging leads to incorrect evaluations when multiple entity types are predicted for the same token.

This ambiguity leads to precision issues: while the model identifies entity spans accurately, it struggles with type classification, particularly for organization entities like football club names that often derive from location names. For example, in the sentence "अब रियल ले जनवरी १६ मा उसै को मैदान मा लेगानेस सँग दोसो लेग खेल्नेछ ।" (*Now Real will play the second leg against Leganes at their own ground on January 16.*), the model might confuse "रियल" (Real) and "लेगानेस" (Leganes) as LOCATION or ORGANIZATION, though the ground truth labels both as ORGANIZATION. As a result, contextual ambiguity contributes to the lower precision scores observed, even though recall remains high as the model consistently identifies the entity spans.

## Impact of Prompt Language

We evaluated the impact of prompt language on NER performance by comparing English and Nepali prompts, using the best-performing English prompts configuration (semantic selection, k=10) for Nepali prompts. Table 6 displays the results for both non-merged and merged approaches.

In the non-merged setting, English prompts outperformed Nepali prompts across all metrics, achieving an F1-micro score of 0.64 compared to 0.57 for Nepali prompts. The recall was particularly higher for English prompts (0.86 vs. 0.66), with English also maintaining slightly better precision.

In the merged setting, the performance gap widened. English prompts remained stable with an F1-micro score of 0.63, while Nepali prompts dropped significantly to 0.50, showing a decline in both precision (0.44) and recall (0.58). This emphasizes the difficulty in maintaining strong performance in Nepali with merged datasets.

These results suggest that, although GPT-4 can process Nepali prompts for NER, English prompts guide the model more effectively, likely due to the model's extensive pre-training on English data. This underscores the need for dedicated pre-training or fine-tuning on low-resource languages like Nepali to improve performance in those languages.

| Lang. | K | Without Merging | | | | Merging | | | |
|---|---|---|---|---|---|---|---|---|---|
| | | Pr. | Re. | $F_m$ | $F_\mu$ | Pr. | Re. | $F_m$ | $F_\mu$ |
| English | 10 | 0.51 | **0.86** | 0.57 | 0.64 | 0.53 | **0.75** | 0.54 | 0.63 |
| Nepali | 10 | 0.50 | **0.66** | 0.50 | 0.57 | 0.44 | **0.58** | 0.41 | 0.50 |

Table 6: Effect of Prompt Language and Merging Strategy on NER Performance. *Note*: Lang., Pr., Re., $F_m$ and $F_\mu$ denote Language, Precision, Recall, F1-macro, and F1-micro, respectively.

## Generative vs. Non-generative LLMs for NER

Table 7 presents a comparison between Generative and Non-generative LLMs across different entity types. For the Generative, we employed semantic selection with k=10. For non-generative, we use the results reported by (Niraula and Chapagain 2022) on the same test set. The non-generative model demonstrated consistently better performance, particularly in precision scores across most entity categories. DATE entities showed the most significant success with the Non-generative approach, achieving an F1-score of 0.91 with balanced precision and recall (both 0.91), compared to the Generative model's F1-score of 0.66.

While the Non-generative AI model consistently outperforms the Generative model across all entity types, the recall of the Generative AI is comparable, and in some cases, even better for specific entities such as EVENT, LOCATION, and PERSON. This suggests that while the Non-generative approach excels in terms of precision and overall performance across all entities, the Generative AI model shows a remarkable ability to identify certain entities, particularly in a low-resource language like Nepali. Despite the lack of fine-tuning and the use of only prompt-based in-context examples, the performance of the Generative model was still relatively strong, demonstrating the potential of generative AI even in challenging linguistic environments. The promising recall results of the Generative model, especially for entities like EVENT and LOCATION, highlight the model's capability to recognize and categorize these entities despite the inherent limitations effectively. Moreover, this data suggests the model's performance could improve significantly through fine-tuning, larger datasets, or expanded prompt examples, potentially enhancing precision and recall across entity types through context-specific optimization.

| Model | Entity | Pre. | Rec. | F$_1$ | Support |
|---|---|---|---|---|---|
| Non-Generative | DATE | 0.91 | **0.91** | 0.91 | 572 |
|  | EVENT | 0.46 | 0.42 | 0.44 | 59 |
|  | LOCATION | 0.85 | 0.80 | 0.82 | 809 |
|  | ORGANIZATION | 0.85 | **0.83** | 0.84 | 715 |
|  | PERSON | 0.90 | 0.85 | 0.88 | 1115 |
| Generative | DATE | 0.55 | 0.81 | 0.66 | 572 |
|  | EVENT | 0.06 | **0.83** | 0.11 | 59 |
|  | LOCATION | 0.57 | **0.87** | 0.69 | 809 |
|  | ORGANIZATION | 0.45 | 0.81 | 0.58 | 715 |
|  | PERSON | 0.79 | **0.90** | 0.84 | 1115 |

Table 7: Comparison of Generative vs. Nongenerative LLMs of the best performing K per named entities. *Note*: Pr., Re., F$_1$ denote Precision, Recall, F1 score respectively.

### Impact of Self-verification

Table 8 demonstrates the self-verification process with an example. When prompted to confirm whether a predicted entity बार्सिलोना is a valid event, the model returned "No," leading us to discard this prediction during evaluation. This illustrates how self-verification can eliminate false positives, enhancing the precision of entity predictions.

> For the sentence: 'रियल शीर्ष स्थान को बार्सिलोना भन्दा १० तथा एट्लेटिको मड्रिड भन्दा ५ अंक ले पछि छ ।'?
> Is the word 'बार्सिलोना' in the given sentence a Event entity? Please answer with Yes or No. No explanation is needed."
> **Model Output:** No

Table 8: Self-verification process for an EVENT entity.

To illustrate the impact of self-verification, we evaluated 100 randomly chosen test instances as running verification on all test instances would have been resource-intensive. Table 9 shows our results on self-verification. It improved precision from 0.50 to 0.66, while recall dropped from 0.85 to 0.71. PERSON entities saw the largest improvement in precision (0.78 to 0.93), with recall remaining stable. ORGANIZATION and LOCATION entities also showed precision gains, while LOCATION recall stayed at 0.88. However, DATE entities experienced a significant recall drop (0.71 to 0.23), despite a slight increase in precision, highlighting the precision-recall trade-off.

These results offer key insights into the use of self-verification for Nepali NER. The approach is most effective for common entity types like PERSON and ORGANIZATION, where the model shows stronger recognition patterns. Self-verification enhances precision by reducing false positives, but the significant recall drop for DATE entities sug-gests the need for entity-specific adjustments in the verification process. The improvement in precision, with minimal recall loss for PERSON and LOCATION entities, indicates that self-verification could be particularly useful for applications prioritizing precision over recall.

| Entity | Without Verification | | | With Verification | | |
|---|---|---|---|---|---|---|
|  | Pre. | Rec. | F1 | Pre. | Rec. | F1 |
| DATE | 0.50 | 0.71 | **0.59** | 0.57 | 0.23 | 0.33 |
| EVENT | 0.07 | 1.00 | 0.13 | 0.10 | 1.00 | **0.18** |
| LOCATION | 0.42 | 0.88 | 0.57 | 0.58 | 0.88 | **0.70** |
| ORGANIZATION | 0.47 | 0.82 | 0.60 | 0.67 | 0.72 | **0.70** |
| PERSON | 0.78 | 0.93 | 0.85 | 0.93 | 0.93 | **0.93** |
| Micro avg | 0.50 | 0.85 | 0.63 | 0.66 | 0.71 | **0.69** |

Table 9: Self-verification on 100 random test examples

### Practical Considerations

Our experiments show that GenAI-based prompting for Nepali NER performs below traditional NER systems (Niraula and Chapagain 2022) (F1-micro: 0.85), with GPT4o achieving F1-micro scores of 0.64 (non-merged) and 0.63 (merged). It is expected that the traditional NER systems learn by utilizing all the training examples (i.e. 13,848) compared to just few examples (i.e. 10) used by the LLM for in-context learning. However, it can provide a strong baseline when limited labeled examples are available especially for common entities (e.g. PERSON) compared to rare entities (e.g. EVENT). Self-verification can boost precision at the cost of recall. Regarding the resource, the total operational cost for the experiment was around $175. The cost for zero-shot was $6, but it increased with more examples, reaching about $33 when using 10 examples for each prompt, reflecting a clear rise as the number of examples grows. Additionally, the lower performance with Nepali prompts (F1-micro: 0.57 vs. 0.64 for English) indicates this approach is better suited for specialized applications than large-scale use.

### Conclusion

This paper evaluated the performance of the state-of-the-art generative LLMs for Nepali NER tasks through various experimental approaches. Our findings reveal that semantic-based example selection significantly outperforms random selection, with F1-micro scores improving from 0.55 to 0.64 in the non-merged setting and from 0.47 to 0.63 in the merged setting. English prompts consistently outperformed Nepali prompts, highlighting the need for improved multilingual capabilities in LLMs. Self-verification boosted precision from 0.50 to 0.66, notably improving PERSON (F1: 0.93) and ORGANIZATION (F1: 0.70) entities, albeit with reduced recall. These results suggest that prompting-based NER is effective for common entities like person names and offers a strong baseline for low-resource settings with limited labeled data. Future work includes instruction fine-tuning to enhance LLM's understanding of Nepali entity patterns and task-specific instructions. Further research should refine self-verification method to boost precision without sacrificing recall, especially for low-performing entities.

# References

Achiam, J.; Adler, S.; Agarwal, S.; Ahmad, L.; Akkaya, I.; Aleman, F. L.; Almeida, D.; Altenschmidt, J.; Altman, S.; Anadkat, S.; et al. 2023. Gpt-4 technical report. *arXiv preprint arXiv:2303.08774*.

Babych, B., and Hartley, A. 2003. Improving machine translation quality with automatic named entity recognition. In *Proceedings of the 7th International EAMT workshop on MT and other language technology tools, Improving MT through other language technology tools, Resource and tools for building MT at EACL 2003*.

Bam, S. B., and Shahi, T. B. 2014. Named entity recognition for nepali text using support vector machines. *Intelligent Information Management* 2014.

Brown, T.; Mann, B.; Ryder, N.; Subbiah, M.; Kaplan, J. D.; Dhariwal, P.; Neelakantan, A.; Shyam, P.; Sastry, G.; Askell, A.; et al. 2020. Language models are few-shot learners. *Advances in neural information processing systems* 33:1877–1901.

Dey, A.; Paul, A.; and Purkayastha, B. S. 2014. Named entity recognition for nepali language: A semi hybrid approach. *International Journal of Engineering and Innovative Technology (IJEIT) Volume* 3:21–25.

He, K.; Mao, R.; Huang, Y.; Gong, T.; Li, C.; and Cambria, E. 2023. Template-free prompting for few-shot named entity recognition via semantic-enhanced contrastive learning. *IEEE transactions on neural networks and learning systems*.

Koirala, P., and Niraula, N. B. 2021. Npvec1: word embeddings for nepali-construction and evaluation. In *Proceedings of the 6th Workshop on Representation Learning for NLP (RepL4NLP-2021)*, 174–184.

Li, J.; Sun, A.; Han, J.; and Li, C. 2020. A survey on deep learning for named entity recognition. *IEEE Transactions on Knowledge and Data Engineering* 34(1):50–70.

Mollá, D.; Van Zaanen, M.; and Smith, D. 2006. Named entity recognition for question answering. In *Proceedings of the Australasian Language Technology Workshop 2006*, 51–58.

Niraula, N., and Chapagain, J. 2022. Named entity recognition for nepali: Data sets and algorithms. In *The International FLAIRS Conference Proceedings*, volume 35.

Niraula, N., and Chapagain, J. 2023. Danfener-named entity recognition in nepali tweets. In *The International FLAIRS Conference Proceedings*, volume 36.

Petroni, F.; Rocktäschel, T.; Lewis, P.; Bakhtin, A.; Wu, Y.; Miller, A. H.; and Riedel, S. 2019. Language models as knowledge bases? *arXiv preprint arXiv:1909.01066*.

Raffel, C.; Shazeer, N.; Roberts, A.; Lee, K.; Narang, S.; Matena, M.; Zhou, Y.; Li, W.; and Liu, P. J. 2020. Exploring the limits of transfer learning with a unified text-to-text transformer. *Journal of machine learning research* 21(140):1–67.

Roller, S.; Dinan, E.; Goyal, N.; Ju, D.; Williamson, M.; Liu, Y.; Xu, J.; Ott, M.; Smith, E. M.; Boureau, Y.-L.; and Weston, J. 2021. Recipes for building an open-domain chatbot. In *Proceedings of the 16th Conference of the European Chapter of the Association for Computational Linguistics: Main Volume*, 300–325. Association for Computational Linguistics.

Sang, E. F., and De Meulder, F. 2003. Introduction to the conll-2003 shared task: Language-independent named entity recognition. *arXiv preprint cs/0306050*.

Singh, O. M.; Padia, A.; and Joshi, A. 2019. Named entity recognition for nepali language. In *2019 IEEE 5th International Conference on Collaboration and Internet Computing (CIC)*, 184–190. IEEE.

Subedi, B.; Regmi, S.; Bal, B. K.; and Acharya, P. 2024. Exploring the potential of large language models (llms) for low-resource languages: A study on named-entity recognition (ner) and part-of-speech (pos) tagging for nepali language. In *Proceedings of the 2024 Joint International Conference on Computational Linguistics, Language Resources and Evaluation (LREC-COLING 2024)*, 6974–6979.

Wang, S.; Sun, X.; Li, X.; Ouyang, R.; Wu, F.; Zhang, T.; Li, J.; and Wang, G. 2023. Gpt-ner: Named entity recognition via large language models. *arXiv preprint arXiv:2304.10428*.

Wei, J.; Wang, X.; Schuurmans, D.; Bosma, M.; Xia, F.; Chi, E.; Le, Q. V.; Zhou, D.; et al. 2022. Chain-of-thought prompting elicits reasoning in large language models. *Advances in neural information processing systems* 35:24824–24837.

Wu, Y.; Schuster, M.; Chen, Z.; Le, Q. V.; Norouzi, M.; Macherey, W.; Krikun, M.; Cao, Y.; Gao, Q.; Macherey, K.; et al. 2016. Google's neural machine translation system: Bridging the gap between human and machine translation. *arXiv preprint arXiv:1609.08144*.

Ye, J.; Xu, N.; Wang, Y.; Zhou, J.; Zhang, Q.; Gui, T.; and Huang, X. 2024. Llm-da: Data augmentation via large language models for few-shot named entity recognition. *arXiv preprint arXiv:2402.14568*.

Zhang, J.; Zhao, Y.; Saleh, M.; and Liu, P. 2020. Pegasus: Pre-training with extracted gap-sentences for abstractive summarization. In *International conference on machine learning*, 11328–11339. PMLR.